\documentclass[letterpaper, 10 pt, conference]{ieeeconf}  % Comment this line out if you need a4paper

\IEEEoverridecommandlockouts                              % This command is only needed if 
\usepackage{graphicx} % for pdf, bitmapped graphics files
\usepackage{amsmath} % assumes amsmath package installed
\usepackage{amssymb}  % assumes amsmath package installed
\usepackage{bm}
\usepackage{url}
\usepackage{subcaption}
\usepackage{color}

\usepackage{tikz}
\usetikzlibrary{positioning}
\usetikzlibrary{arrows.meta}
\usetikzlibrary{plotmarks}
\tikzset{block/.style = {draw}, Connect/.style={-latex,thick}}

\newcommand{\blue}{}

\title{\LARGE \bf
Solving Differential Equations using Physics-Informed Deep Equilibrium Models
}

\author{Bruno M. Pacheco$^{1}$ and Eduardo Camponogara$^{1}$% <-this % stops a space
\thanks{$^{1}$B. Pacheco and E. Camponogara are with the Department of Automation and Systems Engineering, Federal University of Santa Catarina, Brazil, emails: {\tt\footnotesize bruno.m.pacheco@posgrad.ufsc.br}, {\tt\footnotesize eduardo.camponogara@ufsc.br}  }%
}

\begin{document}

\maketitle
\thispagestyle{empty}
\pagestyle{empty}

%%%%%%%%%%%%%%%%%%%%%%%%%%%%%%%%%%%%%%%%%%%%%%%%%%%%%%%%%%%%%%%%%%%%%%%%%%%%%%%%
\begin{abstract}

This paper introduces Physics-Informed Deep Equilibrium Models (PIDEQs) for solving initial value problems (IVPs) of ordinary differential equations (ODEs).
Leveraging recent advancements in deep equilibrium models (DEQs) and physics-informed neural networks (PINNs), PIDEQs combine the implicit output representation of DEQs with physics-informed training techniques.
Our validation of PIDEQs, using the Van der Pol oscillator as a benchmark problem, yielded compelling results, demonstrating their efficiency and effectiveness in solving IVPs.
Our analysis includes key hyperparameter considerations for optimizing PIDEQ performance.
By bridging deep learning and physics-based modeling, this work advances computational techniques for solving IVPs 
 with implications for scientific computing and engineering applications.

\end{abstract}

%%%%%%%%%%%%%%%%%%%%%%%%%%%%%%%%%%%%%%%%%%%%%%%%%%%%%%%%%%%%%%%%%%%%%%%%%%%%%%%%
\section{INTRODUCTION}

The advent of deep learning has revolutionized various industries, demonstrating its prowess in tackling complex problems across domains ranging from image recognition to natural language processing.
Despite the success of deep learning in various industries, applying this technology to solve initial value problems (IVPs) of differential equations presents a formidable challenge, primarily due to the data-driven nature of deep learning models.
Gathering sufficient data from real-life dynamical systems can be prohibitively expensive, necessitating a novel approach to training deep learning models for such tasks.

The work by \cite{RAISSI2019686} introduced the approach of solving IVPs using deep learning models by optimizing a model's dynamics rather than solely its outputs.
This approach allows deep learning models to approximate the dynamics of a system, provided an accurate description of these dynamics is known, typically in the form of differential equations.
Focusing on optimizing the dynamics allows the model to be trained with minimal data, covering only the initial and boundary conditions.

In parallel, \cite{bai_2019_deep} and \cite{ElGhaoui2021} proposed an innovative approach to deep learning by implicitly defining a model's output as a solution to an equilibrium equation.
This methodology results in a model known as a Deep Equilibrium Model (DEQ), which exhibits an infinite-depth network structure with residual connections.
This design offers significant representational power with relatively few parameters, widening the architectural possibilities for deep learning models.

The transition from PINNs to DEQs presents an opportunity to combine the strengths of both approaches.
PINNs excel in incorporating physical laws into the learning process, reducing the dependency on extensive data.
DEQs, with their implicit infinite-depth structure, offer a powerful framework for solving complex problems with fewer parameters.
By integrating these methodologies, we can create a Physics-Informed Deep Equilibrium Model (PIDEQ) that leverages the physics-informed training techniques of PINNs within the robust and efficient framework of DEQs.

This paper explores this integration by studying, implementing, and validating PIDEQs as efficient and accurate solvers for IVPs.
An efficient solver is characterized by its ability to operate with minimal data and computational resources. In contrast, an effective solver can provide accurate solutions across a vast domain of the independent variable.
Our research is guided by three objectives: implementing a DEQ, designing and implementing a PINN training algorithm suitable for DEQs, and evaluating the performance of the physics-informed DEQ in solving IVPs. These objectives form the backbone of our study.
More specifically, our contributions are:
\begin{itemize}
    \item A novel approach for training DEQs for solving IVPs using physics regularization, resulting in a physics-informed deep equilibrium model.

    \item An experimental evaluation of PIDEQs as efficient and effective solvers for ordinary differential equations using the Van der Pol oscillator.
    \item An analysis of the key hyperparameters that must be adjusted to develop PIDEQs effectively.

\end{itemize}

Through these efforts, we aim to provide insights into the suitability of DEQs, a cutting-edge deep learning architecture, for addressing the challenges posed by solving IVPs of ODEs, paving the way for their broader adoption in scientific computing and engineering applications.

%%%%%%%%%%%%%%%%%%
\section{SOLVING IVPs USING DEEP LEARNING}

Solving IVPs of differential equations using deep learning poses unique challenges.
Unlike traditional deep learning tasks where input-output pairs are readily available, in IVPs, both the target function (solution to the IVP) and input-output pairs are unknown or too complex to be directly useful.
This lack of information complicates the training process, as the conventional approach of constructing a dataset of input-output pairs becomes impractical or impossible.

\subsection{Problem Statement}

Consider an IVP of an ODE with boundary conditions.
Given a function $\mathcal{N}: \mathbb{R} \times \mathbb{R}^m \to \mathbb{R}^m$ and initial condition $t_0 \in I \subset \mathbb{R}, \, \bm{y}_0 \subset \mathbb{R}^m$, we aim to find a solution $\bm{\phi}: I \to \mathbb{R}^m$ such that the differential equation \[
\frac{d \bm{\phi}(t)}{d t} = \mathcal{N}\left( t, \bm{\phi}(t) \right), \, \bm{\phi}(t_0) = \bm{y}_0
\] holds on the entirety of the interval $I$.

Our objective is to train a deep learning model $D_{\mathbf{\theta}}: I \to \mathbb{R}^m$, parameterized by $\mathbf{\theta} \in \Theta$, to approximate the solution function $\bm{\phi}$, satisfying the same conditions, i.e., \[
    \frac{d D_{\mathbf{\theta}}(t)}{d t} = \mathcal{N}\left( t, D_{\mathbf{\theta}}(t) \right), \, t \in I \quad \text{and} \quad D_{\mathbf{\theta}}(t_0) = \bm{y}_0
.\]

\subsection{Physics Regularization}

The conventional approach of constructing a dataset for training is inefficient in the context of IVPs, as it does not leverage the known dynamics represented by $\mathcal{N}$.
To address this challenge, \cite{RAISSI2019686} proposed a physics-informed learning approach incorporating known dynamics into the training process.

The key idea is to train the model $D_{\mathbf{\theta}}$ to approximate the solution $\bm{\phi}$ at $t_0$, satisfying the initial condition constraint, and simultaneously train its Jacobian to approximate $\mathcal{N}$, ensuring that the dynamics are captured accurately.

To achieve this, the cost function is defined as
\begin{equation}\label{eq:pinn-loss}
J(\mathbf{\theta}) = J_b(\mathbf{\theta}) + \lambda J_{\mathcal{N}}(\mathbf{\theta})
\end{equation}
with two distinct components weighted by a scalar parameter $\lambda \in \mathbb{R}_+$.
By optimizing this cost function, the model is guided to simultaneously satisfy the initial condition and approximate the dynamics represented by $\mathcal{N}$.

The term $J_b(\mathbf{\theta})$ is introduced to enforce the initial condition constraint.
It is given by \[
    J_b(\mathbf{\theta}) = \| D_{\bm{\theta}}(t_0) - \bm{y}_0 \|_2^2
,\] using an $\ell_2$ norm to penalize the deviation from the initial condition.

The term $J_{\mathcal{N}}(\mathbf{\theta})$ plays a crucial role in ensuring that the model captures the underlying dynamics specified by $\mathcal{N}$.
By regularizing the model's Jacobian, this term forces the model to learn not just the solution at discrete points, but also how the solution evolves according to the differential equations.
This is achieved by minimizing the difference between the model's predicted derivative $\frac{d}{dt} D_{\bm{\theta}}(t)$ and the true dynamics $\mathcal{N}(t, D_{\bm{\theta}}(t))$.
For such, it is defined as \[
    J_{\mathcal{N}}(\mathbf{\theta}) = \sum_{t\sim \mathcal{U}(I)} \left\| \frac{d}{dt} D_{\bm{\theta}}(t) - \mathcal{N}\left( t, D_{\bm{\theta}}(t) \right) \right\|_2^2
,\] where $\mathcal{U}(I)$ is a uniform distribution over the $I$ interval, i.e., the data is uniformly sampled from the domain\footnote{The sampling strategy can be considered a design choice.}.

This approach eliminates the need for explicit knowledge of the target function ($\bm{\phi}$) and allows for efficient training using randomly constructed samples, making deep learning a viable option for solving IVPs.

%%%%%%%%%%%%%%%%%%%%%%%%%%%%%%%%%%%%%%%%%%%%%%%%
\section{DEEP EQUILIBRIUM MODELS}

This section introduces and defines DEQs and presents their combination with physics-informed training in the proposed PIDEQ framework.
We follow the notation proposed by \cite{bai_2019_deep} and explore the foundational concepts and practical considerations that enable the introduction of our proposed approach.

\subsection{Introduction and Definition}

Deep Learning models typically compose simple parametrized functions to capture complex features.
While traditional architectures stack these functions in a sequential manner, in a DEQ, the architecture is based on an infinite stack of the same function.
If this function is well-posed, i.e., it respects a Lipschitz-continuity condition~\cite{ElGhaoui2021}, the infinite stack leads to an equilibrium point that serves as the model's output.

\subsection{Forward Computation}\label{sec:deq-forward}

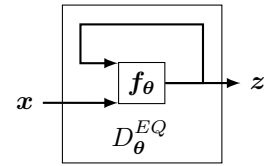
\begin{figure}[h]
    \centering
    \begin{tikzpicture}
        \node (x) {$\bm{x}$};
        \node[style=draw, above right = -0.22 and 1 of x] (f) {$\bm{f}_{\bm{\theta}}$};
        \node[right = of f] (z) {$\bm{z}$};
        % \node[style=draw, right = of x, left = of z, minimum width = 2cm, minimum height = 2cm] (D) {$D_{\bm{\theta}}$};

        \coordinate[above right = 0.5cm and 0.5cm of f](NE);
        \coordinate[above left = 0.5 and 0.5 of f](NW);

        \begin{scope}[every Connect/.style={-latex,thick}]
            \draw [Connect] (f) -- (z);
            \draw [Connect] (f.east)-|(NE)--(NW)|-(f.140);
            \draw [Connect] (x) -- (f.220);
        \end{scope}

        \draw (2.62,1.3) rectangle ++(-2.12,-2.1) node[anchor=south west]{$\quad \, \, \, D^{EQ}_{\bm{\theta}}$};
    \end{tikzpicture}
    \caption{\blue Illustration of the recursion within the equilibrium equation that defines the DEQs.}
    \label{fig:deq-diagram}
\end{figure}

Formally, a DEQ can be defined as the solution to an equilibrium equation.
Let $\bm{f}_{\theta}:\mathbb{R}^n\times\mathbb{R}^{m} \to \mathbb{R}^m$ be the layer function.
Then, the output of a DEQ can be described through the solution $\bm{z}^\star$ of the equilibrium equation \[
    \bm{z} = \bm{f}_{\bm{\theta}}(\bm{x}, \bm{z})
,\] where $\bm{x}$ is the model's input.
\blue
Fig.~\ref{fig:deq-diagram} illustrates the equilibrium equation inside a DEQ.

The simplest way to solve the equilibrium equation that defines the output of a DEQ is by iteratively applying the equilibrium function until convergence.
Given an input $\bm{x}$ and an initial guess $\bm{z}^{[0]}$, the procedure updates the equilibrium guess $\bm{z}^{[i]}$ by \[
    \bm{z}^{[i]} = \bm{f}_{\bm{\theta}}(\bm{x}, \bm{z}^{[i-1]})
\] until $|\bm{z}^{[i]}-\bm{z}^{[i-1]}|$ is sufficiently small.
This approach, known as the simple iteration method, is intuitive but can be slow to converge and is sensitive to the starting point.
Moreover, it only finds equilibrium of functions that are a contraction between the starting point and the equilibrium point~\cite{Sli2003}.
For example, the simple iteration method fails to find the equilibrium of $f(z) = 2z-1$ starting from any point other than the equilibrium itself ($z=1$).

To address these limitations, Newton's method can be used.
It updates the incumbent equilibrium point by \[
    \bm{z}^{[i+1]} = \bm{z}^{[i]} - \left( \frac{d \bm{f}_{\bm{\theta}}(\bm{x},\bm{z}^{[i]})}{d\bm{z}} \right)^{-1} \bm{f}_{\bm{\theta}}(\bm{x},\bm{z}^{[i]})
,\] where $\frac{d}{d\bm{z}} \bm{f}_{\bm{\theta}}(\bm{x},\bm{z}^{[i]})$ represents the Jacobian of $\bm{f}_{\bm{\theta}}$ with respect to $\bm{z}$\footnotemark.
\footnotetext{To avoid the computational burden of inverting the Jacobian matrix, it is common to solve $ \frac{d \bm{f}_{\bm{\theta}}(\bm{x},\bm{z}^{[i]})}{d\bm{z}} \left(\bm{z}^{[i+1]} - \bm{z}^{[i]}\right) = -\bm{f}_{\bm{\theta}}(\bm{x},\bm{z}^{[i]})$ instead.}

Newton's method converges much faster and is applicable to a broader class of functions compared to simple iteration~\cite{Sli2003}.
It offers a faster and more robust alternative by leveraging the Jacobian of $f_{\bm{\theta}}$.
As the deep learning paradigm is to perform gradient-based optimization, the well-posedness of the Jacobian of the equilibrium function is already guaranteed.
In other words, we can benefit from the fast convergence rate and broad applicability of Newton's method.
In fact, most modern root-finding algorithms can be used for computing the equilibrium, such as Anderson acceleration~\cite{Walker2011} and Broyden's Method~\cite{Broyden1965}.

\subsection{Backward Computation}\label{sec:deq-backward}

Computing the gradient of the output of a DEQ with respect to its parameters is not straightforward.
The approach of automatic-differentiation frameworks for deep learning is to backpropagate the loss at training time, differentiating every operation in the forward pass.
Differentiating through modern root-finding algorithms is, at best, an intricate and costly operation.
Even if the forward pass is done through the iterative process, its differentiation would require backpropagating through an unknown number of layers.

Luckily, we can exploit the fact that the output of a DEQ for an equilibrium function $\bm{f}_{\bm{\theta}}(\bm{x},\bm{z})$  defines a \emph{parametrization} of $\bm{z}$ with respect to $\bm{x}$.
This allows us to apply the implicit function theorem to write the Jacobian of a DEQ as
\begin{multline*}
    \frac{d D^{EQ}_{\bm{\theta}}(\bm{x})}{d \bm{\theta}} = \\
    - \left[ \frac{d \bm{f}_{\bm{\theta}}(\bm{x},D^{EQ}_{\bm{\theta}}(\bm{x}))}{d \bm{z}} - I \right]^{-1} \frac{d \bm{f}_{\bm{\theta}}(\bm{x},D^{EQ}_{\bm{\theta}}(\bm{x}))}{d \bm{\theta}}
.\end{multline*}

Note that the Jacobian can be computed \emph{regardless of the operations applied during the forward pass}.
Furthermore, we do not need to compute the entire Jacobian matrix for gradient descent.
\blue
We need to compute the gradient of a loss function $\mathcal{L}$ with respect to the model's parameters.
Such gradient can be written as
\begin{align*}
    \frac{d \mathcal{L}}{d \bm{\theta}} \bigg\rvert_{\hat{y} = D^{EQ}_{\bm{\theta}}(\bm{x})} &= \frac{d \mathcal{L}}{d \hat{y}}\bigg\rvert_{\hat{y}=D^{EQ}_{\bm{\theta}}(\bm{x})} \frac{d D^{EQ}_{\bm{\theta}}}{d \bm{\theta}}\bigg\rvert_{\bm{x}} \\
    &= -\frac{d \mathcal{L}}{d \hat{y}} \left[ \frac{d \bm{f}_{\bm{\theta}}}{d \bm{z}} - I \right]^{-1}\frac{d \bm{f}_{\bm{\theta}}}{d \bm{\theta}}
,\end{align*}
where $\frac{d \mathcal{L}}{d \hat{y}}$ is the gradient of the loss function and $\frac{d D^{EQ}_{\bm{\theta}}}{d \bm{\theta}}$ is the Jacobian of the DEQ with respect to its parameters.
Note that the gradient requires us to compute two vector-matrix products, in which the result of $-\frac{d \mathcal{L}}{d \hat{y}} \left[ \frac{d \bm{f}_{\bm{\theta}}}{d \bm{z}} - I \right]^{-1}$ can be computed through a root-finding algorithm, assuming that the gradient of the loss function is known.
\color{black}
We refer the reader to \cite{bai_2019_deep} and \cite{ElGhaoui2021} for further details.

\subsection{Physics-Informed Deep Equilibrium Model (PIDEQ)}\label{sec:pideq}

A Physics-Informed Deep Equilibrium Model (PIDEQ) extends the principles of physics-informed learning to a DEQ.
In summary, PIDEQs leverage the principles of physics-informed learning to enhance the capabilities of DEQs.
By integrating physics-based constraints, PIDEQs deliver the architectural power of DEQs with robustness and accuracy in modeling physical systems.

The challenge in physics-informing a DEQ lies in optimizing a cost function on its derivatives.
As exposed in Sec.~\ref{sec:deq-backward}, \cite{bai_2019_deep} proposed an efficient method to compute the first derivative of a DEQ with respect to either
its parameters or the input, which, in the context of physics-informed learning, allows us to compute the loss function value. 
However, in computing the gradient of the loss function, we require second derivatives, which pose additional challenges as they imply differentiating the backward pass.
Furthermore, to the best of our knowledge, higher-order derivatives of DEQs have not been investigated.
To address this challenge, we limit ourselves to using solely differentiable operators in the backward pass.
This way, automatic differentiation frameworks can automatically compute the second derivatives.

Finally, it has been shown that DEQs benefit significantly from penalizing the presence of large values in its Jacobian~\cite{pmlr-v139-bai21b}.
This penalization can be achieved by adding the Frobenius norm of the Jacobian of the equilibrium function as a regularization term in the loss function, which was shown to reduce training and inference times.

Therefore, we propose the following loss function for training PIDEQs: \[
    J\left( \bm{\theta} \right) = J_b\left( \bm{\theta} \right) + \lambda J_{\mathcal{N}}\left( \bm{\theta} \right) + \kappa \left\lVert \frac{d \bm{f}_{\bm{\theta}}}{d \bm{z}}\right\rVert_F
.\]
It combines a base loss term ($J_b$) with a physics-informed loss term ($J_{\mathcal{N}}$) and a regularization term based on the Frobenius norm of the Jacobian of the equilibrium function, weighed by a $\kappa \ge 0$ coefficient.

%%%%%%%%%%%%%%%%%%
\section{EXPERIMENTS}

In this section, we conduct a series of experiments to evaluate the capacity of PIDEQs for solving differential equations.
We use the Van der Pol oscillator as our benchmark IVP due to its well-documented complexity and lack of an analytical solution, which makes it an ideal candidate for testing the robustness of numerical solvers.
We compare the PIDEQ approach with physic-informed neural networks (PINNs), a well-established deep learning technique for similar tasks~\cite{RAISSI2019686,antonelo_2021}.

\subsection{Problem Definition}

The Van der Pol oscillator system is chosen as the target ODE for our experiments due to its nonlinear dynamics and the absence of an analytical solution, providing a stringent test for our models.
The system is a classic example in nonlinear dynamics and has been extensively studied, making it a reliable benchmark.

\begin{figure}[h]
    \centering
    \includegraphics[width=0.5\linewidth]{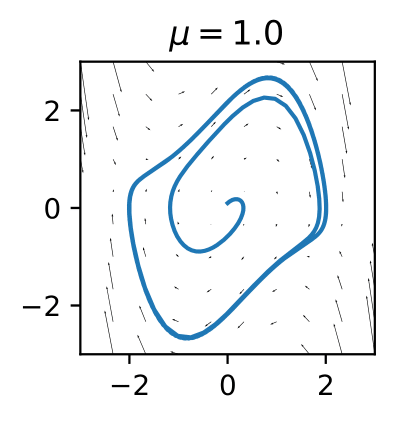}
    \caption{\blue Solution for the Van der Pol oscillator with initial condition $y_1(0)=0,\,y_2(0)=0.1$ and $\mu=1$. The trajectory of the numerical solution is shown in a state-space plot, with $y_2$ in the vertical axis and $y_1$ in the horizontal axis.}\label{fig:vdp-example}
\end{figure}

\blue
The dynamics of the oscillator are described as a system of first-order equations
\begin{align*}
    \frac{d}{dt}\begin{bmatrix} y_1\left( t \right) \\ y_2\left( t \right)  \end{bmatrix} = \begin{bmatrix} 
y_2\left( t \right) \\
\mu\left( 1-y_1\left( t \right) ^2 \right) y_2\left( t \right) - y_1(t)
\end{bmatrix} 
.\end{align*}
We select a value of $\mu=1$ for the dampening coefficient and an initial condition $\bm{y}(0) = (0, 0.1)$, i.e., a small perturbation around the unstable equilibrium at the origin, where $\bm{y}$ represents the two states of the system.
\color{black}
The desired solution is sought over a time horizon of 2 seconds, during which the system is expected to converge to a limit cycle~\cite{Grimshaw2017}, as illustrated in Fig.~\ref{fig:vdp-example}.

\subsubsection{Evaluation Metrics}

Since no analytical solution is available, we assess the performance of the models based on their approximation error compared to a reference solution obtained using the fourth-order Runge-Kutta (RK4) method.
The Integral of the Absolute Error (IAE) is computed over 1000 equally time-spaced steps within the solution interval.
We also consider the computational time required for training and inference as valuable metrics, especially for time-sensitive applications.

\subsection{Training}

In our experiments with the PIDEQ, we use an architecture similar to the one proposed by \cite{ElGhaoui2021}, with an equilibrium function that provides as the output the (element-wise) hyperbolic tangent of an affine combination of its inputs, namely, the time value and the hidden states ($\bm{z}$).
The model's output is a linear combination of the equilibrium vector $\bm{z}^\star$.
Formally, we can write
\begin{equation*}
\begin{split}
    D^{EQ}_{\bm{\theta}}(t) &= C\bm{z}^{\star} \\
    \bm{z}^{\star} &= \bm{f}_{\bm{\theta}}\left( t,\bm{z}^{\star} \right) \\
    \bm{f}_{\bm{\theta}}\left( t,\bm{z} \right) &= \tanh \left( A\bm{z} + t\bm{a} + \bm{b} \right)
\end{split}
\end{equation*}
where the vector parameter is simply a vectorized representation of all coefficients, i.e., $\bm{\theta} = (A,C,\bm{a},\bm{b})${\blue, and the input is $\bm{x} = t$}.

We use the Adam optimizer~\cite{kingma_adam_2014} for training the PIDEQ.
The cost function incorporates regularization terms to enforce physics-informed training, as detailed in Sec.~\ref{sec:pideq}.
Our default hyperparameter configuration\footnote{Determined through early experimentation.} was a learning rate of $10^{-3}$, $\lambda=0.1$, $\kappa=1.0$, and a budget of 50000 epochs.
Our default solver for the forward pass was Anderson acceleration with a tolerance of $10^{-4}$, while the backward pass is always solved iteratively to ensure differentiability.

Nevertheless, the dimension of the hidden states ($|\bm{z}|$), the solver tolerance, and the coefficient for the Jacobian regularization term ($\kappa$) are all considered hyperparameters, and we measure their impacts empirically.
{\blue We also consider as a hyperparameter the choice between Anderson acceleration, Broyden's method, and the simple iteration procedure as the forward pass solver.}

\subsection{Experimental Results}

All experiments reported below were performed on a high-end computer using an RTX 3060 GPU.
Further implementation details can be seen in our code repository\footnote{\url{https://github.com/brunompacheco/pideq}}.
{\blue For every hyperparameter configuration, five models were trained with random initial values for the trainable parameters.}

\subsubsection{Baseline}

\begin{figure}[h]
    \centering
    \includegraphics[width=\linewidth]{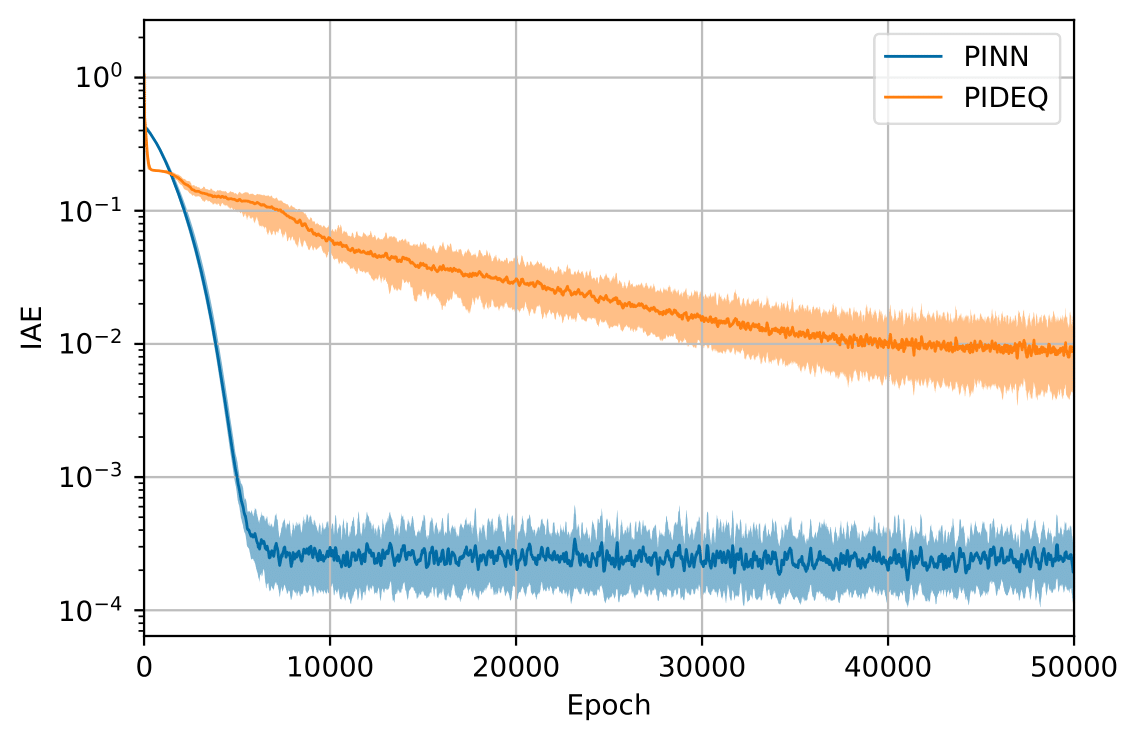}
    \caption{The learning curve for the baseline models trained on the IVP of the Van der Pol oscillator. Solid lines are mean values ($n=5$), and shaded regions represent minimum and maximum values. For a better visualization, a moving average of 100 epochs was taken.}
    \label{fig:baseline-iae}
\end{figure}

Our baseline PINN model follows the architecture proposed in \cite{antonelo_2021}, with four layers of twenty nodes each.
As a DEQ with as many states as there are hidden nodes in a deep feedforward network has \emph{at least} as much representational power~\cite{ElGhaoui2021}, we start with $|\bm{z}| = 80$ hidden states.
The results can be seen in Fig.~\ref{fig:baseline-iae}.
Even though both models achieved low IAE values, the PINN presented better performance in terms of IAE and training time, converging in far fewer epochs.
% with an average of 11~ms per epoch, against 99~ms, besides converging in much fewer epochs.

\subsubsection{Hyperparameter Optimization}

\begin{figure}[h]
    \centering
    \includegraphics[width=\linewidth]{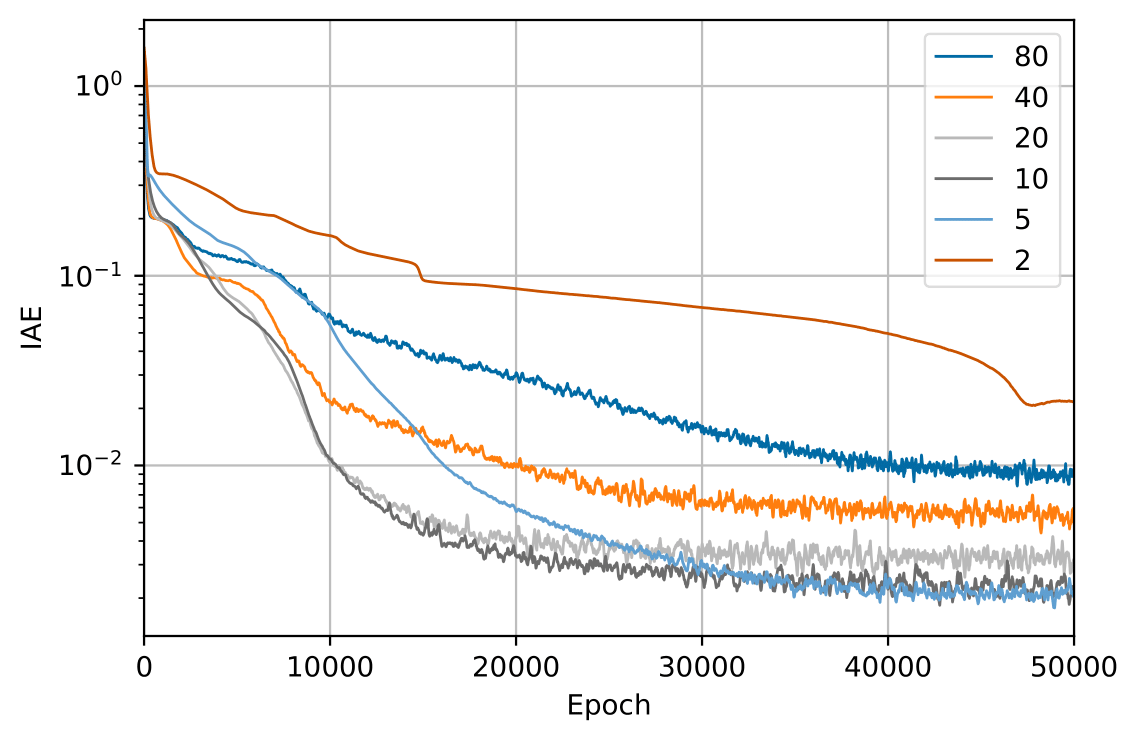}
    \caption{Learning curve of the PIDEQs trained with a varying number of states. The model with 80 hidden states is the same as the baseline PIDEQ from Fig.~\ref{fig:baseline-iae}. Solid lines are mean values ($n=5$). For a better visualization, a moving average of 100 epochs was taken.}
    \label{fig:states-iae}
\end{figure}

A deeper look at the baseline PIDEQ shows that the parameters converged to an $A$ matrix with many null rows.
 This indicates that the model could achieve similar performance with fewer hidden states.
Such intuition proved truthful in our hyperparameter tuning experiments, as shown in Fig.~\ref{fig:states-iae}.
We iteratively halved the number of states until the $A$ matrix had no more empty rows, which happened at $|\bm{z}|=5$ states.
Then, we took one step further, reducing to $|z|=2$ states, but the results indicated that, indeed, 5 was the sweet spot, with faster convergence and lower final IAE.
  The following experiments are performed with PIDEQs with five hidden states.

% \begin{figure}[h]
%     \centering
%     \includegraphics{images/exp_1_matplot.png}
%     \caption{$A$ matrix of the baseline PIDEQ after training. From all models trained, the one with median final performance was used to generate the graphics.}
%     \label{fig:baseline-pideq-A}
% \end{figure}

\begin{figure}[h]
    \centering
    \includegraphics[width=\linewidth]{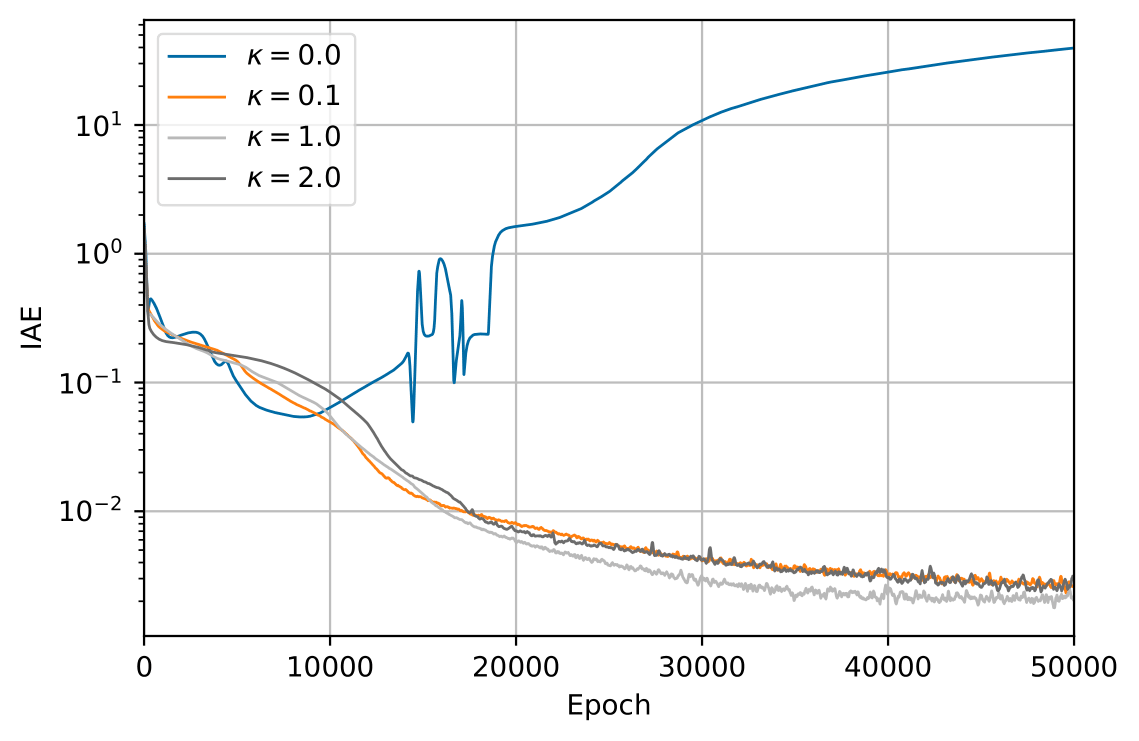}
    \caption{Learning curve of PIDEQs with different $\kappa$ values. Only one model was trained with $\kappa=0$ because the training took over 30 times longer. {\blue For all other scenarios, we present the mean over five runs (parameter initializations)}. For a better visualization, a moving average of 100 epochs was taken.}
    \label{fig:jac-lamb-iae}
\end{figure}

We further investigate the importance of the regularization term on the Jacobian of the equilibrium function.
Even though the Jacobian of the DEQ, in the context of physics-informed training, is directly learned, our experiments indicate that the presence of the regularization term is essential. Otherwise, the training took over 30 times longer and was very unstable.
However, the magnitude of the $\kappa$ coefficient has little impact on the outcomes.
Our results for different values of $\kappa$ (including 0) are illustrated in Fig.~\ref{fig:jac-lamb-iae}.

\begin{figure}[h]
    \centering
    \includegraphics[width=\linewidth]{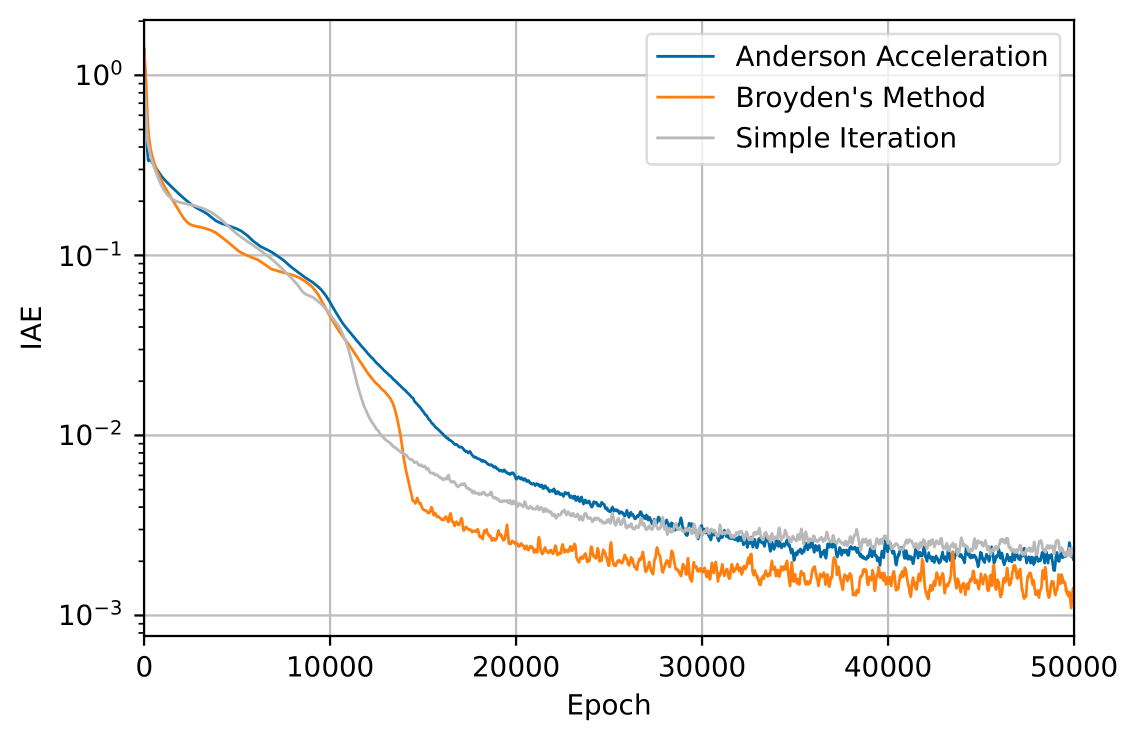}
    \caption{Learning curve of PIDEQs with five states using different solvers for the forward pass. Solid lines are mean values ($n=5$). For a better visualization, a moving average of 100 epochs was taken.}
    \label{fig:solver-iae}
\end{figure}

Anderson acceleration, Broyden's method, and the simple iteration procedure were evaluated as solvers for the forward pass.
As the results illustrated in Fig.~\ref{fig:solver-iae} show, Broyden's method could provide a performance gain, but it came with a significant computational cost, as the epochs took over six times longer in comparison to using Anderson acceleration.
At the same time, using the simple iteration procedure was three times faster than using Anderson acceleration.
However, as discussed in Sec.~\ref{sec:deq-forward}, the theoretical limitation of the simple iterative procedure limits its reliability.
We also evaluated the performance of PIDEQ under different values for the solver tolerance, but no performance gain was observed for values different than our default of $10^{-4}$.

\subsubsection{Results}

\begin{figure}[h]
    \centering
    \begin{subfigure}[t]{.45\linewidth}
	\includegraphics[width=\linewidth]{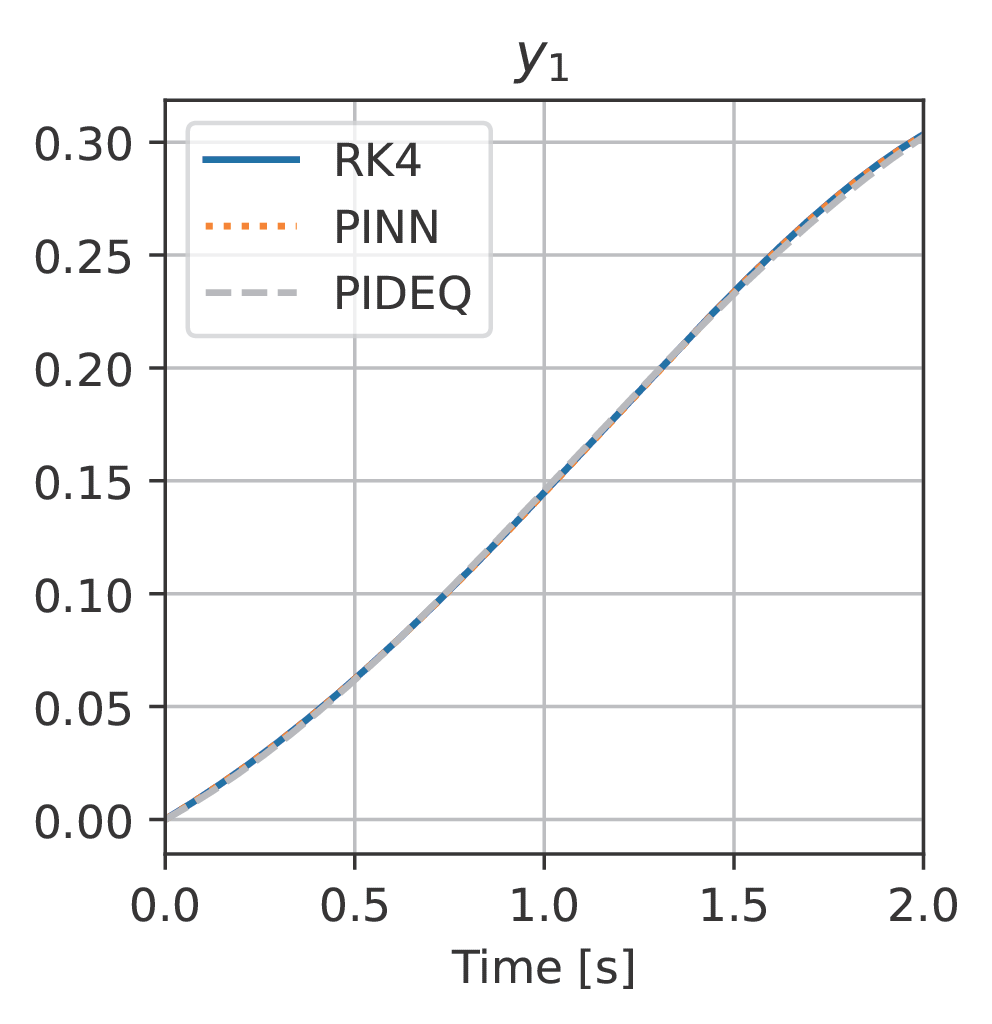}
	\caption{}
    \end{subfigure}
    \begin{subfigure}[t]{.45\linewidth}
	\includegraphics[width=\linewidth]{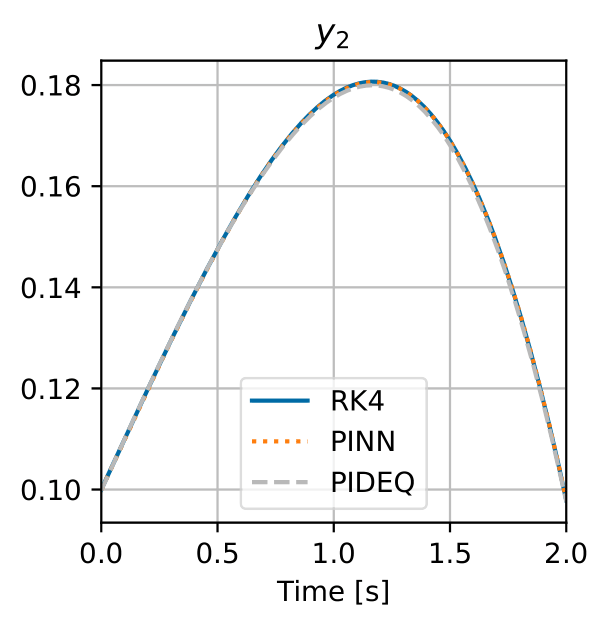}
	\caption{}
    \end{subfigure}
    \caption{Prediction of PINN and PIDEQ in comparison to the reference result from RK4. Both models presented the median performance in the respective experiments.}
    \label{fig:final-vdp}
\end{figure}

\begin{figure}[h]
    \centering
    \includegraphics[width=\linewidth]{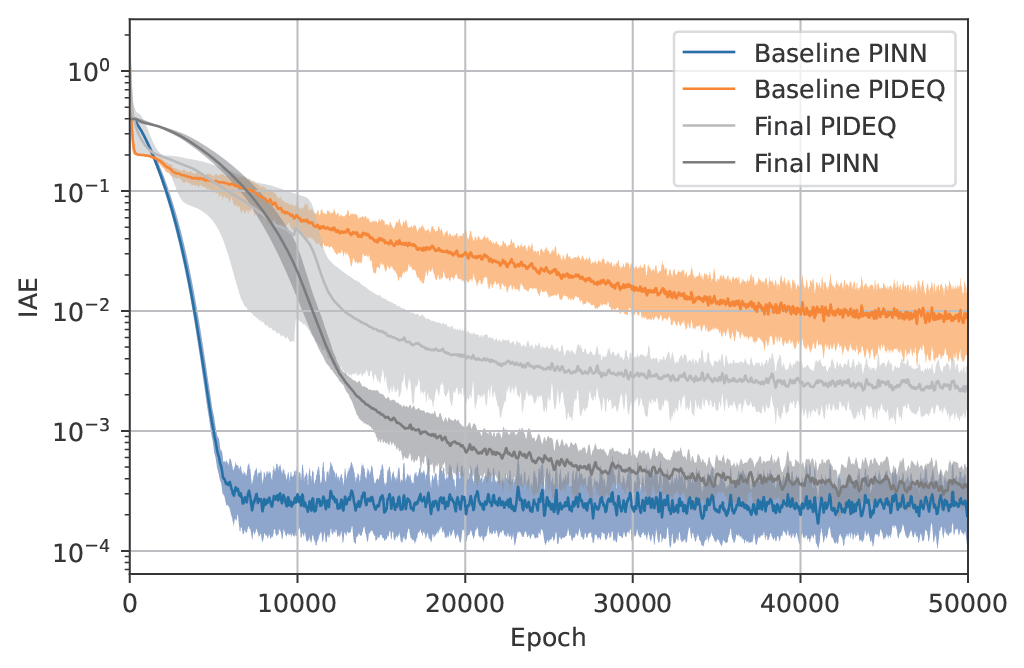}
    \caption{Learning curve of the final models in comparison to the baselines. ``Final PINN'' are the small PINN models, with only 52 parameters. ``Final PIDEQ'' are the PIDEQs after hyperparameter tuning. Solid lines are mean values ($n=5$), and shaded regions represent minimum and maximum values. For a better visualization, a moving average of 100 epochs was taken.}
    \label{fig:final-iae}
\end{figure}

The final PIDEQ, after hyperparameter tuning, achieves comparable results to the baseline PINN models.
In terms of approximation performance, while the baseline PIDEQ achieved an IAE of 0.0082, our best PIDEQ achieved an IAE of 0.0018, and the baseline PINN achieved an IAE of 0.0002, which is seen through the last epochs in Fig.~\ref{fig:final-iae}.
Fig.~\ref{fig:final-vdp} illustrates how small the difference is between the two approaches, showing them in comparison to the solution computed using RK4. 
However, the greatest gain comes from the reduced training time, as the tuned PIDEQ converges much faster than the baseline PIDEQ.

The greatest performance gain came from reducing the number of hidden states, guided by the sparsity of the $A$ matrix.
The smaller model size leads to improved memory efficiency and explainability.
We also trained PINN models with two hidden layers of 5 nodes each, totaling 52 trainable parameters, the same number of parameters as the tuned PIDEQ.
The results are shown in Fig.~\ref{fig:final-iae}.
Note that both final models take around the same number of epochs to converge.

%%%%%%%%%%%%%%%%%%%%%%%%%%%%%%%%%%%%%%%
\section{CONCLUSIONS}

This study explored two innovative approaches to deep learning: Physics-Informed Neural Networks (PINNs) and Deep Equilibrium Models (DEQs).
While PINNs offer efficiency in training models for physical problems, DEQs promise greater representational power with fewer parameters.
We introduced Physics-Informed Deep Equilibrium Models (PIDEQ), uniting the physics-regularization with the infinite-depth architecture.

Our experiments on the Van der Pol oscillator validated PIDEQs' effectiveness in solving IVPs of ODEs, validating our proposed approach.
However, comparisons with PINN models revealed mixed results.
Although PIDEQs demonstrated the ability to solve IVPs, they exhibited slightly higher errors and slower training times compared to PINNs.
This suggests that while DEQ structures offer theoretical advantages in representational power, these benefits may not translate into practical improvements for certain types of problems like the Van der Pol oscillator.

The strengths of PIDEQs lie in their potential for handling more complex problems, leveraging the implicit depth of DEQs.
However, the current implementation's reliance on simple iterative methods for the backward pass can be a limiting factor, especially for more challenging problems and partial differential equations (PDEs).
This limitation points to a significant area for improvement.

Future research should focus on evaluating PIDEQs with more complex problems (e.g., higher-order ODEs and PDEs) to better understand their potential advantages.
Additionally, exploring more sophisticated methods for the backward pass, such as implicit differentiation techniques, could enhance the training efficiency and accuracy of PIDEQs.

In conclusion, while PIDEQs present a promising direction for integrating physics-informed principles with advanced deep-learning architectures, further work is still necessary to fully realize their potential and address their current limitations compared to traditional PINNs.
This future research could pave the way for more robust and versatile models capable of solving a wide range of complex dynamical systems, providing competitive solver alternatives.

\addtolength{\textheight}{-12cm}   % This command serves to balance the column lengths
                                  % on the last page of the document manually. It shortens
                                  % the textheight of the last page by a suitable amount.
                                  % This command does not take effect until the next page
                                  % so it should come on the page before the last. Make
                                  % sure that you do not shorten the textheight too much.

%%%%%%%%%%%%%%%%%%%%%%%%%%%%%%%%%%%%%%%%%%%%%%%%%%%%%%%%%%%%%%%%%%%%%%%%%%%%%%%%

%%%%%%%%%%%%%%%%%%%%%%%%%%%%%%%%%%%%%%%%%%%%%%%%%%%%%%%%%%%%%%%%%%%%%%%%%%%%%%%%

%%%%%%%%%%%%%%%%%%%%%%%%%%%%%%%%%%%%%%%%%%%%%%%%%%%%%%%%%%%%%%%%%%%%%%%%%%%%%%%%
% \section*{APPENDIX}

% Appendixes should appear before the acknowledgment.
\section*{Acknowledgment}

This research was funded in part by Fundação de Amparo à Pesquisa e Inovação do Estado de Santa Catarina (FAPESC) under grant 2021TR2265, CNPq under grants 308624/2021-1 and 402099/2023-0, and CAPES under grant PROEX.

%%%%%%%%%%%%%%%%%%%%%%%%%%%%%%%%%%%%%%%%%%%%%%%%%%%%%%%%%%%%%%%%%%%%%%%%%%%%%%%%

% References are important to the reader; therefore, each citation must be complete and correct. If at all possible, references should be commonly available publications.

\bibliographystyle{ieeetr}

\bibliography{references}

\end{document}